\documentclass{article}
\usepackage{spconf,amsmath,hyperref}

\newcommand{\sys}{\textsc{Almieyar}}

\usepackage{graphicx}
\usepackage[T1]{fontenc}
\usepackage[utf8]{inputenc}

\usepackage{comment}
\usepackage{booktabs}
\usepackage{multirow}
\usepackage{array}
\usepackage{xcolor}
\usepackage{pifont}
\usepackage{subcaption}

\usepackage{tikz}
\usetikzlibrary{shapes.geometric, arrows.meta, positioning, fit, backgrounds}
\usepackage{colortbl}
\usepackage{enumitem}
\usepackage{microtype}

\usepackage{inconsolata}

\tikzset{
  boxBase/.style={draw, rounded corners=3pt, align=center, minimum height=8mm, font=\scriptsize, inner sep=3pt},
  boxNeutral/.style={boxBase, fill=gray!15,   draw=gray!60},
  boxOrange/.style ={boxBase, fill=orange!25, draw=orange!70},
  boxRed/.style    ={boxBase, fill=red!20,    draw=red!60},
  boxBlue/.style   ={boxBase, fill=blue!15,   draw=blue!50},
  pipearrow/.style ={-{Stealth[length=4pt]}, thick, gray!70}
}

\definecolor{IconTeal}{HTML}{0F6E56}
\definecolor{IconTealLight}{HTML}{E1F5EE}
\newcommand{\sysicon}{%
  \mbox{%
  \begin{tikzpicture}[baseline=-0.55ex]
    \draw[IconTeal, line width=0.8pt, fill=IconTealLight, rounded corners=2pt]
      (-1pt, -7.5pt) rectangle (19pt, 7.5pt);
    \fill[IconTeal] ( 1.0pt, -2pt) rectangle ( 2.6pt, 2pt);
    \fill[IconTeal] ( 4.4pt, -3.5pt) rectangle ( 6.0pt, 3.5pt);
    \fill[IconTeal] ( 7.8pt, -5pt) rectangle ( 9.4pt, 5pt);
    \fill[IconTeal] (11.2pt, -3.5pt) rectangle (12.8pt, 3.5pt);
    \fill[IconTeal] (14.6pt, -2pt) rectangle (16.2pt, 2pt);
  \end{tikzpicture}}%
  \,%
}

\title{\sysicon{}\,Almieyar: A Culturally Grounded Benchmark\\for Multi-Dialect Arabic Speech Recognition}

\name{\parbox{\textwidth}{
Omid Ghahroodi\textsuperscript{1},
Anas Madkoor\textsuperscript{1},
Dima Faris Al Saudi\textsuperscript{1},
Fagr Tahir\textsuperscript{2},
Malak Annan\textsuperscript{1},
Talha shahid javad allah rakha\textsuperscript{3},
Omar Al-Busaidi\textsuperscript{4},
Zineb El Kahla\textsuperscript{2},
Iheb Zouari\textsuperscript{1},
Essa Ahmed Abou Jabal\textsuperscript{2},
Ahmed Ezzat\textsuperscript{2},
Hind AL-Merekhi\textsuperscript{1},
Aisha Hamad M A Al-Naimi\textsuperscript{2},
Hadi Wazni\textsuperscript{5},
Bushra Alnajjar\textsuperscript{2},
Omar Amin\textsuperscript{2},
Haya Al-Thani\textsuperscript{1},
Houssam Eddine-Othman LACHEMAT\textsuperscript{1},
Marwa Elwakedy\textsuperscript{6},
Sundus Abdulmalik Al Nahari\textsuperscript{7},
Elahe Zahiri\textsuperscript{1},
Osamah Sarraj\textsuperscript{9},
Raghad Mousa\textsuperscript{10},
Mckeen Assi\textsuperscript{1},
Ahd Al Jumah\textsuperscript{11},
Heyam Salman\textsuperscript{1},
Alhanouf Abdulraqib\textsuperscript{9},
Sara Benoumhani\textsuperscript{12},
Alia Hamwi\textsuperscript{13},
Ayaat Al-Yasseri\textsuperscript{14},
Rim Ibrahim Ghazal\textsuperscript{7},
Lamia Ben hiba\textsuperscript{15},
Mohamed Eltabakh\textsuperscript{1},
Fatima Al-Raisi\textsuperscript{16},
Yassine El Kheir\textsuperscript{17},
Mohammed Abdulrahman\textsuperscript{18},
Hamdy Mubarak\textsuperscript{1},
Ayah Hashem\textsuperscript{3},
Lefkir Meriem\textsuperscript{19},
Ehsaneddin Asgari\textsuperscript{1}
}}

\address{\parbox{\textwidth}{
\textsuperscript{1}QCRI, HBKU;
\textsuperscript{2}Qatar University;
\textsuperscript{3}UDST;
\textsuperscript{4}Algo AI;
\textsuperscript{5}UCL;
\textsuperscript{6}University of Tripoli;
\textsuperscript{7}AUC;
\textsuperscript{9}KAUST;
\textsuperscript{10}CMU-Q;
\textsuperscript{11}KFUPM;
\textsuperscript{12}Alfaisal University;
\textsuperscript{13}Damascus University;
\textsuperscript{14}Princeton University;
\textsuperscript{15}ENSIAS, Mohammed V University;
\textsuperscript{16}Sultan Qaboos University;
\textsuperscript{17}DFKI;
\textsuperscript{18}University of Waterloo;
\textsuperscript{19}USTHB
}}
\begin{document}
%
\maketitle
\begin{abstract}
Arabic speech technology has largely focused on Modern Standard Arabic, leaving the \emph{living} dialects spoken by hundreds of millions under-served. We introduce \sys{}, a culturally grounded ASR benchmark covering 17 Arabic dialects across six families, built entirely from newly recorded speech unseen by existing models. Dialect-community coordinators selected culturally relevant images across 10 topics, and native speakers described them through five structured scenarios, yielding $\approx$50~minutes per dialect (13.7~hours total). We benchmark 12 state-of-the-art ASR systems zero-shot, including GPT-4o-transcribe, Voxtral-Mini-4B, Fanar-STT-LF, Whisper, SeamlessM4T-v2, and wav2vec2-based models. GPT-4o-transcribe achieves the lowest overall WER at $35.0\%$, followed by Voxtral-Mini-4B, Fanar-STT-LF, and Whisper-Large-v3 at $41.1\%$, $45.9\%$, and $49.5\%$, respectively, indicating substantial remaining errors across Arabic dialect communities. Performance varies considerably across dialect groups, with no model performing uniformly best across all groups. WER alone also obscures dialectal ASR behaviour: wav2vec2-based models show large WER/CER gaps, where character-level agreement remains much higher than word-level accuracy, motivating joint WER/CER reporting. \sys{} provides a unified benchmark for culturally grounded Arabic ASR evaluation, including the first published benchmark for Ahwazi Arabic.
\end{abstract}

\section{Introduction}

\label{sec:intro}

Arabic is spoken by over 420 million people across 22 countries \cite{ALQADASI2025103322}, yet its linguistic landscape is far from homogeneous. The gap between Modern Standard Arabic (MSA), used in formal writing and broadcast media, and the diverse spoken dialects is vast, with many varieties mutually unintelligible \cite{zaidan2014arabic}. Critically, \textbf{MSA is not a mother tongue}. The Arabic that people actually \emph{live in}, the Arabic of family gatherings, cultural celebrations, and daily life, is dialectal. Yet ASR research has been overwhelmingly dominated by MSA and a small number of high-resource dialects such as Egyptian \cite{grigoryan2025openautomaticspeechrecognition,obaidah2024newbenchmarkevaluatingautomatic}, leaving many spoken Arabic varieties under-represented in current speech technologies.
This gap affects the accessibility of voice technologies for speakers of dialects such as Moroccan Darija, Sudanese Arabic, and Ahwazi Arabic (the latter with essentially no prior ASR research). Existing multi-dialect benchmarks, including MGB-2 \cite{9003960}, NADI 2025 \cite{talafha-etal-2025-nadi}, and Casablanca \cite{talafha2024casablancadatamodelsmultidialectal}, remain limited both in dialectal coverage, where no prior benchmark distinguishes more than eight varieties (Table~\ref{tab:benchmark_comparison}), and in data provenance, since their speech is sourced from broadcast or social media that models may have seen during training. We address both limitations with \sys{}, a benchmark designed for contamination-free evaluation of Arabic dialect ASR across diverse dialect communities.

\noindent\textbf{Contributions.}
\begin{itemize}[leftmargin=*,topsep=2pt,itemsep=1pt,parsep=0pt]
\item \textbf{Broad dialect coverage.} 17 dialects across six families, including the \emph{first} published ASR benchmark for Ahwazi Arabic, spoken by ${\approx}3$--$5$~million people in Khuzestan, Iran.
\item \textbf{Contamination-free speech.} All recordings are new, elicited specifically for this work; none has appeared in any prior corpus, reducing the risk of evaluation leakage.
\item \textbf{Cultural grounding.} Each utterance describes a community-curated image across 10 thematic topics, eliciting vocabulary and contexts that broadcast-derived corpora may systematically miss.
\item \textbf{Multi-system evaluation.} 12 ASR systems, including open-weight and closed-weight models, are evaluated zero-shot under WER, CER, and MER to analyse model behaviour across diverse Arabic dialects.
\end{itemize}

\section{Related Work}
\label{sec:related}

Large-scale models such as Whisper \cite{pmlr-v202-radford23a} and Conformer-based systems \cite{salhab2025advancing,grigoryan2025openautomaticspeechrecognition} have pushed Arabic ASR WER below~10\%, but primarily on MSA-focused evaluation settings. Standard benchmarks reinforce this imbalance: MGB-2 is drawn from Aljazeera broadcast media and is dialectally mixed but predominantly MSA, while MGB-3 and MGB-5 target Egyptian and Moroccan respectively \cite{9003960,ali2017mgb3,ali2019mgb5}; Common Voice and FLEURS provide Arabic speech resources but have limited representation of Arabic dialect diversity \cite{ardila2019common,conneau2022fleurs}; SADA covers four dialects \cite{alharbi2024sada}; NADI 2025 reaches eight \cite{talafha-etal-2025-nadi}; Casablanca covers eight \cite{talafha2024casablancadatamodelsmultidialectal}. Entire families, including Ahwazi, much of the Gulf, and several Maghrebi varieties, remain absent. Most prior corpora are also drawn from broadcast or YouTube sources, which may overlap with large-scale training data. Where genuinely unseen dialectal speech is evaluated, performance can degrade substantially: \cite{mansour2026sudanese} report $78.8\%$ zero-shot WER on Sudanese Arabic. Culturally grounded benchmarking has gained traction in NLP \cite{zahraei2025ialignedwhommena,abootorabi-etal-2026-almieyar} but has not been widely applied to Arabic ASR; \sys{} fills this gap. Table~\ref{tab:benchmark_comparison} positions it against prior Arabic ASR benchmarks: it combines broad dialectal coverage with newly generated, culturally grounded speech designed for contamination-free evaluation.

\begin{table}[t]
  \centering
  \small
  \setlength{\tabcolsep}{3pt}
  \begin{tabular}{lcc}
    \toprule
    \textbf{Benchmark} & \textbf{Dialects} & \textbf{Hours} \\
    \midrule
    MGB-2 \cite{9003960} & MSA-dom. & 1,200 \\
    MGB-3 \cite{ali2017mgb3} & 1 (EGY) & 16 \\
    MGB-5 \cite{ali2019mgb5} & 1 (MOR) & 13 \\
    Common Voice \cite{ardila2019common} & Arabic varieties & varies \\
    FLEURS \cite{conneau2022fleurs} & Arabic varieties & varies \\
    SADA \cite{alharbi2024sada} & 4 & 668 \\
    NADI 2025 \cite{talafha-etal-2025-nadi} & 8 & $\approx$20 \\
    Casablanca \cite{talafha2024casablancadatamodelsmultidialectal} & 8 & -- \\
    \midrule
    \textbf{\sys{} (ours)} & \textbf{17} & \textbf{13.7} \\
    \bottomrule
  \end{tabular}
  \caption{Comparison of Arabic ASR benchmarks. ``MSA-dom.'': benchmark data are primarily MSA despite containing some dialectal speech.}
  \label{tab:benchmark_comparison}
\end{table}

\section{The \sys{} Benchmark}
\label{sec:method}

\begin{figure*}[t]
\centering
\includegraphics[width=0.72\textwidth, trim=0 2.2cm 0 0, clip]{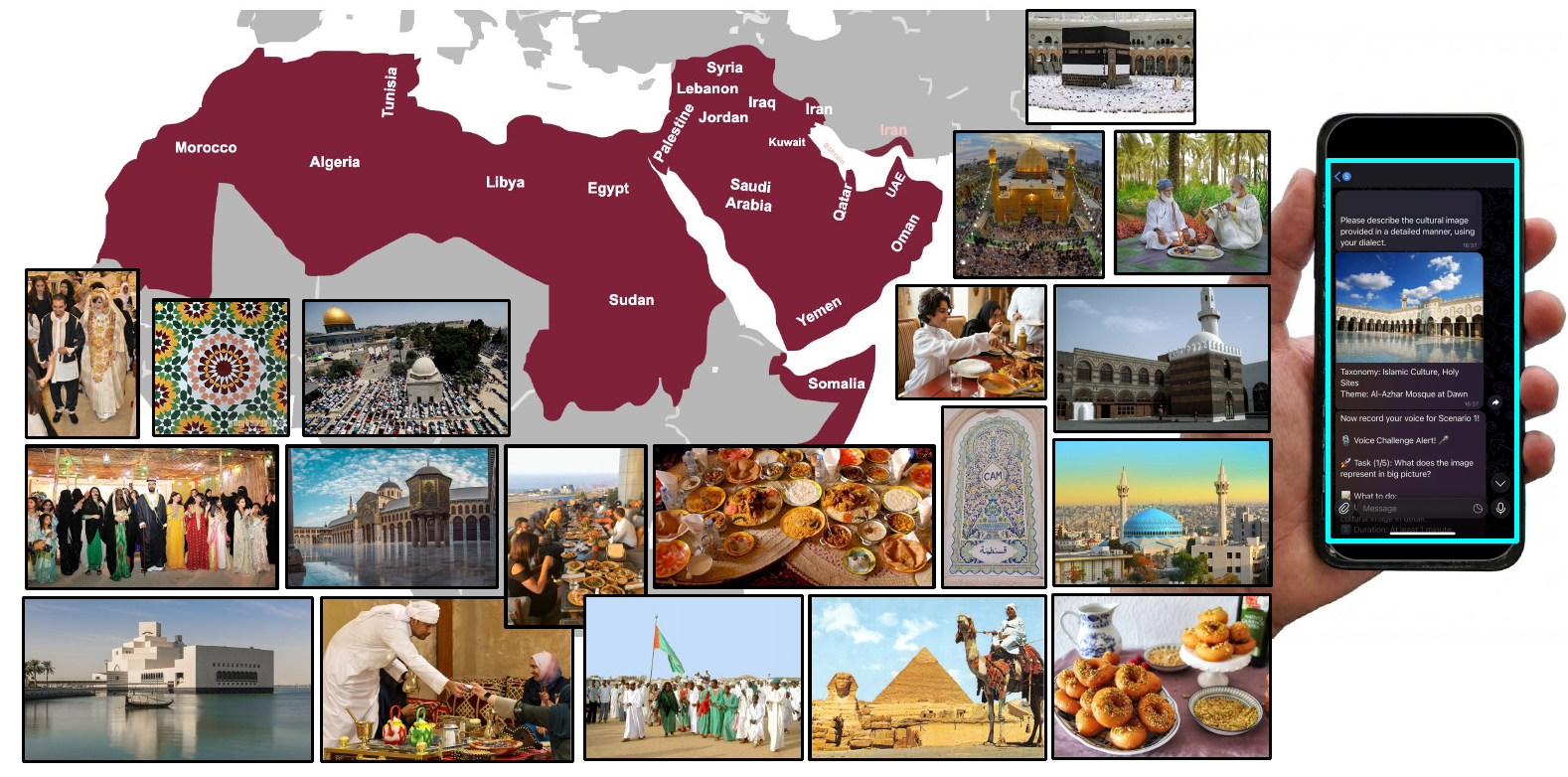}
\vspace{2mm}
\resizebox{0.72\textwidth}{!}{%
\begin{tikzpicture}[node distance=5mm and 4mm]
  \node[boxNeutral, text width=14mm] (img)  {Image +\\Scenario};
  \node[boxOrange,  right=of img,  text width=12mm] (rec)  {Speaker\\Recording};
  \node[boxOrange,  right=of rec,  text width=14mm] (asr)  {Auto ASR\\Transcript};
  \node[boxOrange,  right=of asr,  text width=14mm] (edit) {Speaker\\Correction};
  \node[boxRed,     right=of edit, text width=18mm] (rev)  {Coordinator\\Review};
  \node[boxBlue,    right=of rev,  text width=18mm] (out)  {Verified\\Dialect ASR Data};
  \draw[pipearrow] (img) -- (rec);
  \draw[pipearrow] (rec) -- (asr);
  \draw[pipearrow] (asr) -- (edit);
  \draw[pipearrow] (edit) -- (rev);
  \draw[pipearrow] (rev) -- (out);
\end{tikzpicture}}
\caption{
  \textbf{\sys{} benchmark construction pipeline.} Dialect speakers describe culturally selected images across five structured scenarios using a Telegram-based collection interface. Automatic transcription is used only to assist speaker correction; final references are verified through speaker correction and coordinator review.
}
\label{fig:pipeline}
\end{figure*}
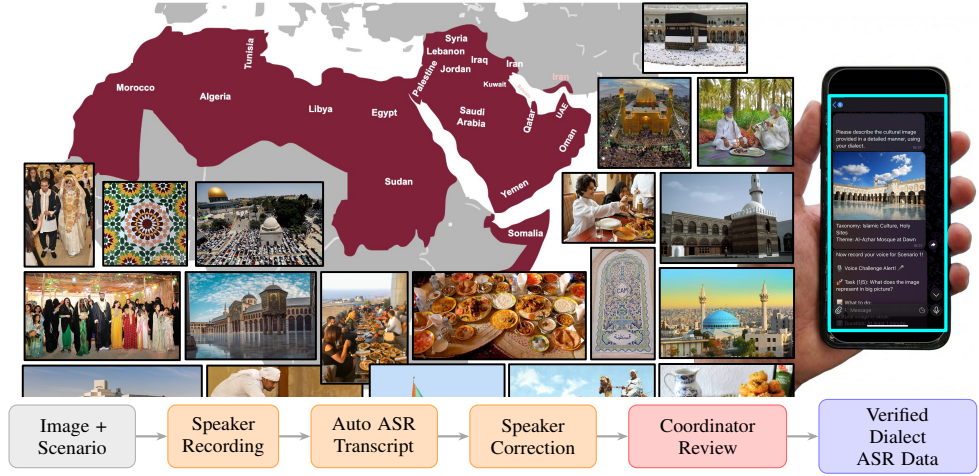

\begin{table*}[t]
  \centering
  \setlength{\tabcolsep}{2pt}
  \small
  \begin{tabular}{l cccccc @{\hspace{6pt}} |c| @{\hspace{10pt}} cccccc @{\hspace{6pt}} |c|}
      \toprule
      & \multicolumn{7}{c}{\textbf{WER (\%)}} & \multicolumn{7}{c}{\textbf{CER (\%)}} \\
      \cmidrule(lr){2-8} \cmidrule(lr){9-15}
      \textbf{Model}
        & \textbf{Gulf} & \textbf{Lev.} & \textbf{Mag.} & \textbf{Iraqi} & \textbf{Egy.} & \textbf{Sud.} & \textbf{Ovr.}
        & \textbf{Gulf} & \textbf{Lev.} & \textbf{Mag.} & \textbf{Iraqi} & \textbf{Egy.} & \textbf{Sud.} & \textbf{Ovr.} \\
      \midrule
            GPT-4o-transcribe
        & \cellcolor[RGB]{82,176,81} 36.4
        & \cellcolor[RGB]{64,168,76} 33.6
        & \cellcolor[RGB]{76,173,79} 34.9
        & \cellcolor[RGB]{45,159,70} \textbf{28.3}
        & \cellcolor[RGB]{237,247,174} 49.7
        & \cellcolor[RGB]{58,165,74} 30.6
        & \cellcolor[RGB]{26,150,65} \textbf{34.9}
        & \cellcolor[RGB]{76,173,79} 19.3
        & \cellcolor[RGB]{64,168,76} 14.5
        & \cellcolor[RGB]{70,171,77} 15.7
        & \cellcolor[RGB]{26,150,65} \textbf{9.4}
        & \cellcolor[RGB]{254,229,161} 33.9
        & \cellcolor[RGB]{26,150,65} \textbf{10.2}
        & \cellcolor[RGB]{26,150,65} \textbf{16.5} \\
      Voxtral-Mini-4B
        & \cellcolor[RGB]{26,150,65}  \textbf{35.7}
        & \cellcolor[RGB]{145,206,99} 47.1
        & \cellcolor[RGB]{58,165,74}  \textbf{38.8}
        & \cellcolor[RGB]{117,193,91} \textbf{44.4}
        & \cellcolor[RGB]{237,247,174} 59.9
        & \cellcolor[RGB]{26,150,65}  \textbf{35.8}
        & \cellcolor[RGB]{82,176,81}  \textbf{41.1}
        & \cellcolor[RGB]{64,168,76}  \textbf{17.9}
        & \cellcolor[RGB]{169,218,109} 26.3
        & \cellcolor[RGB]{45,159,70}  \textbf{16.5}
        & \cellcolor[RGB]{118,194,92}  22.1
        & \cellcolor[RGB]{254,229,161} 40.2
        & \cellcolor[RGB]{26,150,65}  \textbf{15.0}
        & \cellcolor[RGB]{98,184,86}  \textbf{20.6} \\
      Fanar-STT-LF
        & \cellcolor[RGB]{93,182,84}   42.2
        & \cellcolor[RGB]{126,197,94}  \textbf{45.3}
        & \cellcolor[RGB]{174,220,114} 50.4
        & \cellcolor[RGB]{139,204,98}  46.6
        & \cellcolor[RGB]{158,213,103} 48.4
        & \cellcolor[RGB]{89,180,83}   41.8
        & \cellcolor[RGB]{132,201,96}  45.9
        & \cellcolor[RGB]{93,182,84}   20.2
        & \cellcolor[RGB]{92,181,84}  \textbf{20.1}
        & \cellcolor[RGB]{129,199,95}  23.0
        & \cellcolor[RGB]{92,181,84}  \textbf{20.1}
        & \cellcolor[RGB]{179,222,118} 27.4
        & \cellcolor[RGB]{40,156,69}   16.1
        & \cellcolor[RGB]{104,187,87}  21.0 \\
      Whisper-Large-v3
        & \cellcolor[RGB]{141,205,98}  46.8
        & \cellcolor[RGB]{195,229,133} 53.5
        & \cellcolor[RGB]{171,219,111} 50.0
        & \cellcolor[RGB]{149,208,101} 47.5
        & \cellcolor[RGB]{217,238,155} 56.8
        & \cellcolor[RGB]{114,192,90}  44.2
        & \cellcolor[RGB]{168,218,108} 49.5
        & \cellcolor[RGB]{188,226,127} 28.6
        & \cellcolor[RGB]{219,239,157} 32.4
        & \cellcolor[RGB]{185,225,124} 28.2
        & \cellcolor[RGB]{161,214,104} 25.5
        & \cellcolor[RGB]{250,252,186} 36.1
        & \cellcolor[RGB]{135,202,96}  23.4
        & \cellcolor[RGB]{191,228,130} 29.0 \\
      SeamlessM4T-v2
        & \cellcolor[RGB]{186,225,125} 52.2
        & \cellcolor[RGB]{253,211,140} 69.7
        & \cellcolor[RGB]{208,235,146} 55.6
        & \cellcolor[RGB]{172,219,112} 50.2
        & \cellcolor[RGB]{110,190,89}  \textbf{43.9}
        & \cellcolor[RGB]{212,236,150} 56.1
        & \cellcolor[RGB]{215,238,153} 56.6
        & \cellcolor[RGB]{238,247,175} 34.7
        & \cellcolor[RGB]{237,114,69}  52.0
        & \cellcolor[RGB]{254,252,187} 37.1
        & \cellcolor[RGB]{207,234,145} 30.9
        & \cellcolor[RGB]{168,218,108} \textbf{26.2}
        & \cellcolor[RGB]{254,248,183} 37.6
        & \cellcolor[RGB]{254,241,175} 38.5 \\
      Whisper-Medium
        & \cellcolor[RGB]{217,238,154} 56.8
        & \cellcolor[RGB]{239,248,176} 60.2
        & \cellcolor[RGB]{229,244,166} 58.6
        & \cellcolor[RGB]{221,240,159} 57.5
        & \cellcolor[RGB]{254,239,172} 65.1
        & \cellcolor[RGB]{222,241,160} 57.6
        & \cellcolor[RGB]{228,243,165} 58.5
        & \cellcolor[RGB]{233,245,170} 34.1
        & \cellcolor[RGB]{231,245,168} 33.9
        & \cellcolor[RGB]{218,239,155} 32.2
        & \cellcolor[RGB]{225,242,162} 33.1
        & \cellcolor[RGB]{254,218,149} 41.6
        & \cellcolor[RGB]{209,235,147} 31.2
        & \cellcolor[RGB]{229,243,166} 33.6 \\
      Whisper-Small
        & \cellcolor[RGB]{254,244,178} 64.2
        & \cellcolor[RGB]{251,168,94}  76.4
        & \cellcolor[RGB]{253,200,128} 71.4
        & \cellcolor[RGB]{254,217,147} 68.7
        & \cellcolor[RGB]{253,178,101} 75.2
        & \cellcolor[RGB]{244,250,181} 60.9
        & \cellcolor[RGB]{253,211,140} 69.7
        & \cellcolor[RGB]{254,232,164} 39.8
        & \cellcolor[RGB]{244,142,82}  49.9
        & \cellcolor[RGB]{254,221,152} 41.2
        & \cellcolor[RGB]{253,214,143} 42.2
        & \cellcolor[RGB]{243,138,80}  50.2
        & \cellcolor[RGB]{234,246,171} 34.2
        & \cellcolor[RGB]{253,209,138} 42.8 \\
      Moonshine-AR
        & \cellcolor[RGB]{253,208,136} 70.2
        & \cellcolor[RGB]{254,222,153} 67.8
        & \cellcolor[RGB]{247,153,87}  77.6
        & \cellcolor[RGB]{254,218,149} 68.4
        & \cellcolor[RGB]{235,107,66}  81.8
        & \cellcolor[RGB]{254,249,184} 63.4
        & \cellcolor[RGB]{253,199,126} 71.7
        & \cellcolor[RGB]{250,163,92}  48.4
        & \cellcolor[RGB]{253,198,125} 44.3
        & \cellcolor[RGB]{234,102,63}  52.8
        & \cellcolor[RGB]{253,194,120} 44.8
        & \cellcolor[RGB]{215,25,28}   58.5
        & \cellcolor[RGB]{254,226,158} 40.5
        & \cellcolor[RGB]{250,163,92}  48.4 \\
      Wav2Vec2-53-AR
        & \cellcolor[RGB]{250,163,91}  76.8
        & \cellcolor[RGB]{237,114,69}  81.2
        & \cellcolor[RGB]{236,108,66}  81.7
        & \cellcolor[RGB]{253,184,109} 74.1
        & \cellcolor[RGB]{228,77,52}   84.5
        & \cellcolor[RGB]{253,181,105} 74.6
        & \cellcolor[RGB]{243,138,80}  79.1
        & \cellcolor[RGB]{190,227,129} 28.8
        & \cellcolor[RGB]{208,235,146} 31.0
        & \cellcolor[RGB]{200,231,139} 30.1
        & \cellcolor[RGB]{141,205,98}  23.9
        & \cellcolor[RGB]{254,217,147} 41.8
        & \cellcolor[RGB]{152,210,102} 24.8
        & \cellcolor[RGB]{195,229,134} 29.4 \\
      Wav2Vec2-XLSR-AR
        & \cellcolor[RGB]{243,134,78}  79.4
        & \cellcolor[RGB]{231,89,57}   83.4
        & \cellcolor[RGB]{228,77,52}   84.5
        & \cellcolor[RGB]{245,143,82}  78.6
        & \cellcolor[RGB]{218,38,34}   88.0
        & \cellcolor[RGB]{247,153,87}  77.7
        & \cellcolor[RGB]{235,106,65}  81.9
        & \cellcolor[RGB]{214,237,152} 31.8
        & \cellcolor[RGB]{229,244,166} 33.6
        & \cellcolor[RGB]{221,240,159} 32.7
        & \cellcolor[RGB]{174,220,113} 26.9
        & \cellcolor[RGB]{253,204,131} 43.6
        & \cellcolor[RGB]{178,222,117} 27.3
        & \cellcolor[RGB]{217,239,155} 32.1 \\
      Wav2Vec2-XLSR-v2
        & \cellcolor[RGB]{233,98,61}   82.7
        & \cellcolor[RGB]{226,70,48}   85.2
        & \cellcolor[RGB]{219,43,36}   87.5
        & \cellcolor[RGB]{232,92,59}   83.2
        & \cellcolor[RGB]{215,25,28}   89.2
        & \cellcolor[RGB]{239,121,72}  80.5
        & \cellcolor[RGB]{227,74,50}   84.8
        & \cellcolor[RGB]{220,240,158} 32.5
        & \cellcolor[RGB]{232,245,169} 33.9
        & \cellcolor[RGB]{235,246,172} 34.3
        & \cellcolor[RGB]{188,226,127} 28.6
        & \cellcolor[RGB]{253,195,122} 44.7
        & \cellcolor[RGB]{188,226,127} 28.6
        & \cellcolor[RGB]{226,242,163} 33.2 \\
      Sinai-STT
        & \cellcolor[RGB]{224,63,45}   85.8
        & \cellcolor[RGB]{219,44,37}   87.5
        & \cellcolor[RGB]{220,46,38}   87.3
        & \cellcolor[RGB]{237,115,69}  81.1
        & \cellcolor[RGB]{221,48,39}   87.1
        & \cellcolor[RGB]{249,158,89}  77.2
        & \cellcolor[RGB]{225,65,46}   85.6
        & \cellcolor[RGB]{254,215,145} 42.0
        & \cellcolor[RGB]{254,219,150} 41.4
        & \cellcolor[RGB]{254,240,174} 38.7
        & \cellcolor[RGB]{220,240,157} 32.5
        & \cellcolor[RGB]{253,179,103} 46.9
        & \cellcolor[RGB]{182,224,122} 27.9
        & \cellcolor[RGB]{254,235,168} 39.3 \\
      \bottomrule
    \end{tabular}
  \caption{Family-level WER (\%) and CER (\%) per model on \sys{}. Cells are colour-coded green (low) to red (high). Best per column in \textbf{bold}. \textbf{Gulf}: Bahraini, Omani, Qatari, Saudi, Yemeni; \textbf{Lev.}: Levantine (Jordanian, Lebanese, Palestinian, Syrian); \textbf{Mag.}: Maghrebi (Algerian, Libyan, Moroccan, Tunisian); \textbf{Iraqi}: Iraqi \& Ahwazi; \textbf{Egy.}: Egyptian; \textbf{Sud.}: Sudanese; \textbf{Ovr.}: Overall average.}
  \label{tab:full_wer_cer}
\end{table*}

\begin{table}[t]
  \centering
  \scriptsize
  \setlength{\tabcolsep}{2pt}
  \begin{tabular}{l ccccccc}
    \toprule
    \textbf{Model} & \textbf{Gulf} & \textbf{Lev.} & \textbf{Mag.} & \textbf{Irq.} & \textbf{Egy.} & \textbf{Sud.} & \textbf{Ovr.} \\
    \midrule
    Voxtral-Mini-4B
      & \cellcolor[RGB]{26,150,65}  \textbf{33.4}
      & \cellcolor[RGB]{106,188,88} \textbf{41.4}
      & \cellcolor[RGB]{66,169,76}  \textbf{37.5}
      & \cellcolor[RGB]{100,185,86} \textbf{40.8}
      & \cellcolor[RGB]{180,223,119} 49.6
      & \cellcolor[RGB]{26,150,65}  \textbf{33.4}
      & \cellcolor[RGB]{70,171,77}  \textbf{37.8} \\
    Fanar-STT-LF
      & \cellcolor[RGB]{110,190,89} 41.8
      & \cellcolor[RGB]{136,203,97} 44.4
      & \cellcolor[RGB]{182,223,121} 49.9
      & \cellcolor[RGB]{156,212,103} 46.4
      & \cellcolor[RGB]{145,207,100} 45.3
      & \cellcolor[RGB]{104,187,88} 41.3
      & \cellcolor[RGB]{145,207,99} 45.3 \\
    Whisper-Large-v3
      & \cellcolor[RGB]{126,198,94} 43.4
      & \cellcolor[RGB]{180,223,120} 49.7
      & \cellcolor[RGB]{158,213,103} 46.6
      & \cellcolor[RGB]{135,202,97} 44.3
      & \cellcolor[RGB]{176,221,115} 48.9
      & \cellcolor[RGB]{100,185,86} 40.8
      & \cellcolor[RGB]{151,209,101} 45.9 \\
    SeamlessM4T-v2
      & \cellcolor[RGB]{190,227,129} 51.2
      & \cellcolor[RGB]{253,207,136} 69.4
      & \cellcolor[RGB]{217,238,154} 55.4
      & \cellcolor[RGB]{180,223,120} 49.7
      & \cellcolor[RGB]{97,184,85}  \textbf{40.5}
      & \cellcolor[RGB]{215,237,152} 55.1
      & \cellcolor[RGB]{220,240,158} 55.9 \\
    Whisper-Medium
      & \cellcolor[RGB]{196,229,134} 52.1
      & \cellcolor[RGB]{216,238,154} 55.3
      & \cellcolor[RGB]{210,235,148} 54.3
      & \cellcolor[RGB]{207,234,145} 53.8
      & \cellcolor[RGB]{240,248,177} 59.0
      & \cellcolor[RGB]{195,229,134} 52.0
      & \cellcolor[RGB]{207,234,145} 53.9 \\
    Whisper-Small
      & \cellcolor[RGB]{254,253,189} 61.5
      & \cellcolor[RGB]{253,193,119} 71.9
      & \cellcolor[RGB]{254,219,150} 67.4
      & \cellcolor[RGB]{254,227,158} 66.1
      & \cellcolor[RGB]{253,191,117} 72.2
      & \cellcolor[RGB]{234,246,171} 58.1
      & \cellcolor[RGB]{254,226,158} 66.2 \\
    Moonshine-AR
      & \cellcolor[RGB]{254,217,147} 67.7
      & \cellcolor[RGB]{254,221,151} 67.1
      & \cellcolor[RGB]{250,163,92}  76.3
      & \cellcolor[RGB]{253,214,143} 68.3
      & \cellcolor[RGB]{241,128,75}  79.5
      & \cellcolor[RGB]{254,252,188} 61.7
      & \cellcolor[RGB]{253,203,130} 70.3 \\
    Wav2Vec2-53-AR
      & \cellcolor[RGB]{249,159,90}  76.7
      & \cellcolor[RGB]{237,112,68}  81.0
      & \cellcolor[RGB]{235,105,65}  81.7
      & \cellcolor[RGB]{253,180,105} 74.1
      & \cellcolor[RGB]{228,77,52}   84.3
      & \cellcolor[RGB]{253,182,106} 73.8
      & \cellcolor[RGB]{243,135,79}  78.9 \\
    Wav2Vec2-XLSR-AR
      & \cellcolor[RGB]{242,133,78}  79.0
      & \cellcolor[RGB]{231,88,57}   83.2
      & \cellcolor[RGB]{228,76,51}   84.4
      & \cellcolor[RGB]{244,140,81}  78.4
      & \cellcolor[RGB]{219,41,35}   87.6
      & \cellcolor[RGB]{249,160,90}  76.6
      & \cellcolor[RGB]{235,106,65}  81.6 \\
    Wav2Vec2-XLSR-v2
      & \cellcolor[RGB]{234,100,63}  82.1
      & \cellcolor[RGB]{227,72,49}   84.8
      & \cellcolor[RGB]{219,43,36}   87.5
      & \cellcolor[RGB]{232,92,59}   82.9
      & \cellcolor[RGB]{215,25,28}   89.2
      & \cellcolor[RGB]{240,124,74}  79.9
      & \cellcolor[RGB]{227,75,51}   84.5 \\
    Sinai-STT
      & \cellcolor[RGB]{224,61,45}   85.8
      & \cellcolor[RGB]{219,43,36}   87.4
      & \cellcolor[RGB]{220,46,37}   87.2
      & \cellcolor[RGB]{236,111,67}  81.1
      & \cellcolor[RGB]{220,47,38}   87.1
      & \cellcolor[RGB]{247,152,87}  77.2
      & \cellcolor[RGB]{224,63,45}   85.6 \\
    \bottomrule
  \end{tabular}
  \caption{MER (\%) per model and dialect family on \sys{}. Cells are colour-coded green (low) to red (high). Best per column in \textbf{bold}. Family abbreviations match Table~\ref{tab:full_wer_cer}.}
  \label{tab:full_mer}
\end{table}

\textbf{Dialect coverage.} The 17 target dialects across six families are represented by \emph{dialect-community coordinators}: native speakers with linguistic awareness who contributed cultural expertise and joined the project as co-authors. Most speakers were young adults (approximately 20--30 years), with gender distributions varying across communities due to contributor availability.

\noindent \textbf{Cultural image selection.} Coordinators selected culturally representative images across \textbf{10 topics}: food, clothing, religious practices, historical architecture, festivals, crafts, nature, family gatherings, markets, and sports. Images were selected to reflect authentic community contexts.

\noindent \textbf{Five-scenario elicitation and pipeline.} Speakers described each image for $\approx$30--60~s using five structured scenarios covering cultural context, subjects, background, colours/mood, and interpretation. The collection yielded $\approx$50~minutes per dialect and \textbf{13.7~hours} overall. A Telegram-based pipeline (Figure~\ref{fig:pipeline}) uses automatic transcription only for speaker correction, followed by coordinator review to verify dialect authenticity and flag code-switching.

\section{Evaluation Framework}
\label{sec:eval}

\textbf{Models.} We benchmark 12 ASR systems zero-shot: \textbf{GPT-4o-transcribe}, \textbf{Voxtral-Mini-4B} \cite{mistral2025voxtral}, \textbf{Fanar-STT-LF}, \textbf{Whisper-Large-v3}, \textbf{Whisper-Medium}, \textbf{Whisper-Small} \cite{pmlr-v202-radford23a}, \textbf{SeamlessM4T-v2} \cite{seamless2023}, \textbf{Moonshine-AR} \cite{moonshine}, \textbf{Wav2Vec2-53-AR}, \textbf{Wav2Vec2-XLSR-AR}, \textbf{Wav2Vec2-XLSR-v2} \cite{baevski2020wav2vec}, and \textbf{Sinai-STT}\footnote{\url{https://huggingface.co/bakrianoo/sinai-voice-ar-stt}}. Since all data are newly collected, results measure out-of-domain generalisation. Models were evaluated using standard inference settings: Whisper in Arabic transcription mode, wav2vec2 systems with greedy CTC decoding, and other systems with default generation procedures.

\noindent\textbf{Metrics.} We evaluate against manually reviewed references after identical Arabic normalization, including removal of diacritics and tatweel, unification of Alef/Hamza variants, normalization of Ta Marbuta and Alef Maqsura, and collapsing punctuation and whitespace variation. We report WER and CER for word- and character-level recognition. CER is particularly informative for Arabic due to morphological complexity and cliticisation. MER, which normalizes word errors by total word events, is additionally reported to capture cases where WER is affected by hypothesis length.

\section{Results and Discussion}
\label{sec:results}
\textbf{Top tier and WER/CER dissociation.}
Table~\ref{tab:full_wer_cer} reports WER and CER for all 12 systems across six dialect families. \textsc{GPT-4o-transcribe} achieves the lowest overall error ($34.9\%$ WER, $16.5\%$ CER), followed by \textsc{Voxtral-Mini-4B} ($41.1\%/20.6\%$), \textsc{Fanar-STT-LF} ($45.9\%/21.0\%$), and \textsc{Whisper-Large-v3} ($49.5\%/29.0\%$). Wav2Vec2-based systems and \textsc{Sinai-STT} remain substantially behind ($79$--$86\%$ WER). Their large WER/CER gaps indicate partial character-level agreement despite incorrect word recognition, with errors mainly involving deletions and substitutions of weak letters and short function morphemes. Dialect-to-MSA substitutions occur but are not the dominant failure pattern, highlighting the importance of reporting both WER and CER.

\noindent\textbf{Family-level variation.}
ALMIEYAR evaluates model behaviour across dialect communities rather than ranking dialect difficulty. Error rates vary across families and systems due to differences in model coverage, training exposure, and benchmark content. Substantial variation also exists within families: dialect-level gaps reach $16.8$ points in Maghrebi and $12.8$ points in Gulf, showing that family averages can hide dialect-specific behaviour.

\noindent\textbf{Gulf (4.0~h).}\quad Gulf achieves relatively strong results, with \textsc{Voxtral} and \textsc{GPT-4o-transcribe} reaching $35.7\%$ and $36.4\%$ WER, respectively. Performance varies across Bahraini, Omani, Qatari, Saudi, and Yemeni Arabic, motivating evaluation beyond family-level aggregation.

\noindent\textbf{Levantine (3.0~h).}\quad \textsc{GPT-4o-transcribe} achieves the lowest Levantine WER ($33.6\%$), followed by \textsc{Fanar-STT-LF} ($45.3\%$) and \textsc{Voxtral} ($47.1\%$). \textsc{SeamlessM4T-v2} shows a large WER/CER discrepancy ($69.7\%$ WER, $52.0\%$ CER), indicating substantial character-level confusion.

\noindent\textbf{Maghrebi (3.8~h).}\quad Maghrebi remains challenging for several systems, with notable variation among dialects. \textsc{GPT-4o-transcribe} achieves $34.9\%$ average WER, while weaker systems show substantially higher error rates. Wav2Vec2 models again exhibit large WER/CER gaps, reflecting partial character overlap despite word-level errors.

\noindent\textbf{Iraqi and Ahwazi (1.6~h).}\quad This low-resource setting shows strong performance for leading models: \textsc{GPT-4o-transcribe} achieves $28.3\%$ WER, followed by \textsc{Voxtral} ($44.4\%$) and \textsc{Fanar-STT-LF} ($46.6\%$). The combined family contains variation between Iraqi and Ahwazi Arabic, with Ahwazi providing a previously uncovered evaluation setting.

\noindent\textbf{Egyptian (0.7~h).}\quad Egyptian Arabic shows high error rates for several systems. \textsc{GPT-4o-transcribe}, \textsc{Voxtral}, and \textsc{Moonshine-AR} achieve $49.7\%$, $59.9\%$, and $81.8\%$ WER, respectively, while \textsc{SeamlessM4T-v2} achieves $43.9\%$. 

\noindent\textbf{Sudanese (0.7~h).}\quad \textsc{GPT-4o-transcribe} achieves $30.6\%$ WER, followed by \textsc{Voxtral} at $35.8\%$ WER and $15.0\%$ CER. Lower-performing systems exceed $74\%$ WER, showing continued challenges for under-represented dialects.

\noindent\textbf{Model comparison across dialects.}
\textsc{GPT-4o-transcribe} provides the strongest overall result, while other systems show relative strengths across specific dialect groups. These differences demonstrate that aggregate scores alone do not fully capture dialectal ASR behaviour and motivate evaluation across diverse communities.

\section{Match Error Rate Results}
\label{app:mer}

Table~\ref{tab:full_mer} reports MER across evaluated models and six dialect families. MER normalises word-level errors by total word events, making it more conservative than WER when hypotheses contain more words than the references. Results are broadly consistent with Table~\ref{tab:full_wer_cer}. The Wav2Vec2 family and \textsc{Sinai-STT} remain among the highest-error systems, with their large WER/CER differences indicating substantial character-level overlap despite incorrect word-level predictions.

\section{Conclusions}
\label{sec:conclusion}

We introduced \sys{}, a culturally grounded and contamination free Arabic ASR benchmark covering 17 dialects across six families, including the first published Ahwazi benchmark. The benchmark is constructed through community-guided image descriptions using entirely newly recorded speech. Across 12 zero-shot systems, including both open-weight and closed-weight models, current ASR systems still exhibit substantial errors on diverse Arabic dialect communities, with GPT-4o-transcribe achieving the strongest overall performance at $34.9\%$ WER. The variation across systems and dialect groups highlights the importance of evaluating ASR models across diverse communities rather than relying on a single aggregate score. The WER/CER analysis further shows that different architectures exhibit different failure patterns, demonstrating the need for multi-metric evaluation of Arabic dialect ASR.





\section{Limitations}
\sys{} represents a meaningful step toward equitable dialectal Arabic ASR evaluation, but several limitations remain.
\textbf{(i) Single elicitation modality.} The benchmark is built entirely from image-description speech. While effective at surfacing culturally grounded vocabulary, it does not capture other registers such as conversational dialogue, read speech, broadcast monologue, or domain-specific discourse (medical, legal, technical).
\textbf{(ii) Scale per dialect.} With $\approx$50 minutes per dialect, performance estimates carry non-trivial variance for the lower-resource families (e.g., Ahwazi, Sudanese, Iraqi); statistical comparisons between similarly performing models on a single dialect should be interpreted as indicative rather than definitive.
\textbf{(iii) Discrete dialect categories.} Arabic dialects form a continuum rather than a partition; our 17-way labelling is a working approximation. Speaker-level metadata (age, gender, urban/rural background, education) is recorded but not used as a stratification variable in headline results.
\textbf{(iv) Zero-shot only.} We evaluate models out-of-the-box; performance after dialect-specific fine-tuning is left to future work.
\textbf{(v) Reference quality.} Coordinator review reduces but does not eliminate transcription ambiguity, especially around code-switching boundaries and dialect-internal orthographic conventions where no single normative spelling exists.

\section{Ethical Considerations}

\textbf{Contributor roles and authorship.}
We distinguish two roles in \sys{}'s construction, with the ethical implications of each handled differently:
\begin{itemize}[leftmargin=*,topsep=2pt,itemsep=1pt,parsep=0pt]
\item \emph{Dialect-community coordinators} carried out substantive intellectual work: curating culturally representative image sets for their dialect, recruiting and supervising speakers, calibrating elicitation prompts, performing final transcription review, and adjudicating dialect-internal orthographic conventions. Following the standard CRediT / ICMJE criteria for authorship (substantial contribution, drafting/review, and final approval), each coordinator is listed as a co-author of this paper. 
\item \emph{Native-speaker participants} contributed bounded one-time recordings ($\approx$50~minutes per speaker). Their contribution falls below the threshold for authorship under CRediT, so it is recognised through a fair monetary compensation paid for their time, plus a named acknowledgement (with explicit consent) in the released datasheet. We followed local minimum-wage benchmarks and approved IRB compensation rates in each contributor country.
\end{itemize}

\textbf{Informed consent.}
All speakers gave written informed consent for use of their voice recordings and transcriptions in research and in the public release of \sys{}. Consent forms were provided in the speaker's dialect (oral readback where literacy was uneven) and explicitly enumerated the licence, the planned redistribution, and the right to withdraw before publication.

\textbf{Personally identifying information (PII).}
We retain only the dialect label and a study-internal speaker ID with the released audio; speaker names, contact details, demographic metadata at individual granularity, and any incidental PII surfaced during recording (e.g.,~mentions of family members) were redacted or pseudonymised by the coordinators before release.

\textbf{Cultural representation.}
Image selection and transcript review reflect community judgement about cultural authenticity rather than top-down decisions made by the core authors. Where dialects span national boundaries (e.g.,~Iraqi/Ahwazi, Levantine across four states), multiple coordinators were consulted; disagreements were resolved by majority vote of coordinators native to the disputed sub-region.

\textbf{Intended use and misuse.}
\sys{} is intended for the \emph{evaluation} of Arabic ASR systems; it is not a training corpus, and we explicitly discourage its repackaging as one to avoid contaminating future benchmarks. Released ASR outputs reflect model behaviour as of evaluation time and should not be used to make individual-level claims about speakers.

\textbf{Release and datasheet.}
The dataset, all evaluation scripts, model-output transcripts, and per-utterance metric reports will be released under a permissive licence, alongside a datasheet documenting collection methodology, processing pipeline, coordinator and speaker compensation rates, and known biases (e.g.,~within-dialect speaker-demographic skews).

\appendix

\section{Five-Scenario Elicitation Instructions}
\label{app:scenarios}

Instructions were provided to coordinators in both English and their dialect's written Arabic.

\noindent\textbf{Scenario 1, Cultural Context.}\quad Describe the cultural image in detail using your dialect. Focus on: (a) the big picture and what the image represents; (b) where the scene is set; (c) the specific tradition, festival, or activity depicted; (d) historical or cultural significance; (e) any clues about time of day or season.

\noindent\textbf{Scenario 2, Central Subjects.}\quad Describe the primary figures or objects in the image. Cover: (a) who or what dominates the image; (b) their placement and orientation; (c) specific colours, textures, or notable details; (d) how multiple subjects relate to each other.

\noindent\textbf{Scenario 3, Background and Environment.}\quad Describe the setting and surroundings. Address: (a) what is visible in the background; (b) architecture, natural elements, or decorative features; (c) how the background complements or contrasts the central subjects; (d) patterns, textures, or recurring elements.

\noindent\textbf{Scenario 4, Colours, Textures, and Mood.}\quad Describe the sensory and emotional qualities. Include: (a) dominant colours and their locations; (b) textures of objects or surfaces; (c) cultural symbolism of specific colours; (d) how lighting and shadows create emotional tone.

\noindent\textbf{Scenario 5, Interpretive Meaning.}\quad Describe the deeper significance of the image. Address: (a) the cultural narrative or message; (b) the emotional impact on you as a viewer; (c) connection to cultural, historical, or spiritual roots; (d) why this scene is meaningful in a modern context.

\section{Telegram Bot Collection Interface}
\label{app:bot}

The Telegram bot (Figure~\ref{fig:pipeline}) guided each participant through: \textbf{(1)}~display image with scenario instructions in the speaker's dialect; \textbf{(2)}~record voice message (30--60~s); \textbf{(3)}~auto-transcribe via Google Cloud ASR; \textbf{(4)}~display transcription in editable field for in-line correction; \textbf{(5)}~submit and repeat for remaining scenarios. Recordings were stored with metadata: speaker ID, dialect, image ID, scenario number, timestamp, and duration. Coordinator review used a separate web interface with line-by-line comparison of original and corrected transcriptions.

\begin{table*}[t]
  \centering
  \scriptsize
  \setlength{\tabcolsep}{2.2pt}
  \begin{tabular}{l ccccccccccc @{\hspace{10pt}} ccccccccccc}
    \toprule
    & \multicolumn{11}{c}{\textbf{WER (\%)}} & \multicolumn{11}{c}{\textbf{CER (\%)}} \\
    \cmidrule(lr){2-12} \cmidrule(lr){13-23}
    \textbf{Dialect} & Vox & Fan & W-L & Sea & W-M & W-S & Moo & W53 & XLS & XLv2 & Sin & Vox & Fan & W-L & Sea & W-M & W-S & Moo & W53 & XLS & XLv2 & Sin \\
    \midrule
    \multicolumn{23}{l}{\textit{Gulf}} \\
    \quad Bahraini & 43 & 41 & 61 & 54 & 70 & 76 & 73 & 78 & 80 & 82 & 85 & 23 & 17 & 44 & 30 & 48 & 58 & 50 & 27 & 30 & 30 & 40 \\
    \quad Omani & 30 & 41 & 51 & 76 & 64 & 67 & 67 & 78 & 81 & 83 & 89 & 11 & 22 & 28 & 65 & 38 & 44 & 42 & 29 & 34 & 32 & 50 \\
    \quad Qatari & 35 & 42 & 41 & 38 & 53 & 58 & 77 & 76 & 79 & 82 & 83 & 17 & 18 & 23 & 17 & 29 & 31 & 55 & 27 & 29 & 30 & 37 \\
    \quad Saudi & 25 & 38 & 36 & 42 & 45 & 59 & 52 & 73 & 77 & 82 & 83 & 11 & 19 & 20 & 24 & 26 & 33 & 34 & 28 & 30 & 32 & 37 \\
    \quad Yemeni & 39 & 52 & 52 & 73 & 59 & 65 & 76 & 81 & 82 & 85 & 89 & 18 & 28 & 32 & 58 & 34 & 38 & 54 & 35 & 38 & 39 & 51 \\
    \addlinespace[2pt]
    \multicolumn{23}{l}{\textit{Levantine}} \\
    \quad Jordanian & 35 & 39 & 46 & 74 & 54 & 72 & 57 & 80 & 80 & 85 & 88 & 18 & 16 & 27 & 57 & 32 & 48 & 35 & 28 & 29 & 31 & 44 \\
    \quad Lebanese & 49 & 59 & 50 & 45 & 62 & 70 & 63 & 90 & 94 & 92 & 88 & 20 & 26 & 20 & 16 & 29 & 28 & 28 & 38 & 40 & 42 & 37 \\
    \quad Palestinian & 53 & 49 & 58 & 77 & 64 & 80 & 80 & 81 & 83 & 85 & 87 & 29 & 23 & 36 & 60 & 36 & 56 & 57 & 33 & 36 & 36 & 41 \\
    \quad Syrian & 38 & 44 & 54 & 64 & 60 & 76 & 69 & 81 & 83 & 84 & 87 & 17 & 19 & 35 & 46 & 35 & 51 & 46 & 30 & 33 & 33 & 41 \\
    \addlinespace[2pt]
    \multicolumn{23}{l}{\textit{Maghrebi}} \\
    \quad Algerian & 51 & 58 & 58 & 66 & 65 & 77 & 66 & 86 & 88 & 91 & 88 & 26 & 29 & 36 & 45 & 37 & 44 & 43 & 35 & 37 & 39 & 42 \\
    \quad Libyan & 27 & 40 & 41 & 45 & 49 & 61 & 57 & 78 & 80 & 84 & 84 & 9 & 16 & 22 & 27 & 27 & 36 & 30 & 29 & 32 & 33 & 40 \\
    \quad Moroccan & 48 & 53 & 71 & 73 & 69 & 82 & 84 & 84 & 86 & 89 & 92 & 22 & 26 & 42 & 53 & 43 & 50 & 56 & 33 & 35 & 37 & 42 \\
    \quad Tunisian & 36 & 52 & 45 & 49 & 57 & 70 & 91 & 81 & 85 & 88 & 87 & 13 & 23 & 23 & 31 & 28 & 40 & 67 & 28 & 30 & 32 & 36 \\
    \addlinespace[2pt]
    \multicolumn{23}{l}{\textit{Iraqi}} \\
    \quad Ahwazi & 43 & 57 & 57 & 43 & 69 & 77 & 78 & 79 & 84 & 87 & 86 & 18 & 30 & 30 & 19 & 40 & 48 & 51 & 29 & 34 & 35 & 41 \\
    \quad Iraqi Arabic & 35 & 36 & 39 & 58 & 46 & 59 & 60 & 70 & 74 & 80 & 75 & 16 & 11 & 22 & 43 & 27 & 36 & 40 & 19 & 21 & 23 & 24 \\
    \bottomrule
  \end{tabular}
  \caption{Per-dialect WER (\%) and CER (\%) on \sys{} for the four multi-dialect families, to the nearest whole percent. Model abbreviations, in the column order of Table~\ref{tab:full_wer_cer}: \textbf{Vox}~= Voxtral-Mini-4B, \textbf{Fan}~= Fanar-STT-LF, \textbf{W-L}~= Whisper-Large-v3, \textbf{Sea}~= SeamlessM4T-v2, \textbf{W-M}~= Whisper-Medium, \textbf{W-S}~= Whisper-Small, \textbf{Moo}~= Moonshine-AR, \textbf{W53}~= Wav2Vec2-53-AR, \textbf{XLS}~= Wav2Vec2-XLSR-AR, \textbf{XLv2}~= Wav2Vec2-XLSR-v2, \textbf{Sin}~= Sinai-STT.}
  \label{tab:per_dialect}
\end{table*}

\vfill\pagebreak




\bibliographystyle{IEEEbib}
\bibliography{strings,refs}

\end{document}